%% file: neurips_2025.tex
\documentclass{article}

\usepackage[preprint]{neurips_2025}
\usepackage{amsmath}
\usepackage{graphicx}
\usepackage{multirow}
\usepackage{colortbl} 
\usepackage[utf8]{inputenc}
\definecolor{lightblue}{RGB}{230, 235, 255}  
\usepackage[acronym]{glossaries}
\makeglossaries
\newacronym{bnnm}{BNNM}{Bidirectional Nearest Neighbor Merging}
\newacronym{ldsd}{LDSD}{Limited Directional Sequential Dependence}
\newacronym{hsp}{HSP}{Hidden States Protection}
\newacronym{stm}{STM}{Sequential Token Merging}
\usepackage[utf8]{inputenc} % allow utf-8 input
\usepackage[T1]{fontenc}    % use 8-bit T1 fonts
\usepackage{hyperref}       % hyperlinks
\usepackage{url}            % simple URL typesetting
\usepackage{booktabs}       % professional-quality tables
\usepackage{amsfonts}       % blackboard math symbols
\usepackage{nicefrac}       % compact symbols for 1/2, etc.
\usepackage{microtype}      % microtypography
\usepackage{xcolor}         % colors

\title{Sequential Token Merging: Revisiting Hidden States}

\author{%
    Yan Wen\textsuperscript{1,2},~
    Peng Ye\textsuperscript{3,4},~
    Lin Zhang\textsuperscript{1},~
    Baopu Li\textsuperscript{5},\\
    \textbf{Jiakang Yuan\textsuperscript{1}},~
    \textbf{Yaoxin Yang\textsuperscript{1}},~
    \textbf{Tao Chen\textsuperscript{1,2}\textsuperscript{\dag}}\\
    \textsuperscript{1}College of Future Information Technology, Fudan University \\
    \textsuperscript{2}Shanghai Innovation Institute\\
    \textsuperscript{3}Shanghai Artificial Intelligence Laboratory\\
    \textsuperscript{4}The Chinese University of Hong Kong~~
    \textsuperscript{5}Independent Researcher\\
    \texttt{21307130037@m.fudan.edu.cn}\quad
    \texttt{eetchen@fudan.edu.cn}\\
}
\begin{document}

\maketitle

\renewcommand{\thefootnote}{\fnsymbol{footnote}}
\footnotetext[1]{\textsuperscript{\dag} Corresponding author.}

\begin{abstract}
Vision Mambas (ViMs) achieve remarkable success with sub-quadratic complexity, but their efficiency remains constrained by quadratic token scaling with image resolution. While existing methods address token redundancy, they overlook ViMs' intrinsic Limited Directional Sequential Dependence (LDSD)—a critical information flow mechanism revealed in our analysis. We further identify Mamba’s selective scan enables gradual information aggregation in hidden states. Based on these insights, we propose Sequential Token Merging (STM), featuring: 1)~Bidirectional nearest neighbor merging to preserve sequential dependencies through symmetric spatial aggregation, and 2)~Hidden states protection to stabilize the hidden states around the class token. STM strategically leverages Mamba's layer-wise loss convergence to convert temporal forgetfulness into stability. Experiments demonstrate STM's superiority: 1.0\% accuracy drop for ViM-Ti at 20\% token reduction, and only 1.4\% degradation for ViM-S at 40\% reduction. Our method achieves state-of-the-art efficiency with minimal complexity, while providing new insights into state-space model dynamics. Codes will be released soon.

\end{abstract}

\section{Introduction}
Vision Mambas (ViMs) are generic vision backbones based on Mamba, a hardware-accelerated  State Space Model (SSM)~\cite{gu2021efficiently,zhu_vision_2024}. ViMs are famous for their sub-quadratic complexity~\cite{han_demystify_2024,liu_vision_2024} compared to Vision Transformers (ViTs)~\cite{dosovitskiy2020image,liu2022swin,touvron2021training,zhang2023hivit} in computer vision. Similar to ViTs, larger ViMs encounter deployment difficulties stemming from substantial memory usage and latency limitations \cite{chiang_quamba_2024}.  Many efforts have been devoted to more efficient and accelerated ViMs for visual tasks~\cite{cho_ptq4vm_2024,lei_dvmsr_2024,pei_efficientvmamba_2025,ramachandran_ouromamba_2025,shi_post-training_2025}. Token reduction is a common technique in ViTs to balance the tradeoff between computational efficiency and model accuracy~\cite{kim_token_2024,pan2021ia,rao2021dynamicvit,renggli2022learning,yuan2021tokens}. Given its success in ViTs, token reduction has shown to be a useful method to enhance the efficiency of ViMs because the token lengths are usually fixed and independent of the model architectures. \cite{shi_faster_2024,zhan_exploring_2024} center around the reduction of unnecessary tokens by transferring techniques from ViTs to ViMs, such as inter-token similarity metrics and attention-derived importance scores. However, these methods might not follow the mechanisms in ViMs, leading to an unsatisfactory accuracy and deeply relying on fine-tuning or retraining for restoration.

The above issue may be attributed to an unproven assumption in ViMs that some tokens can be removed because they contribute little to the output. However, since Mamba’s hidden states selectively propagate information, such detection-and-removal strategies can lead to unpredictable and uncontrollable loss in hidden states. Figure \ref{fig1} visualizes heatmaps of hidden states and attention scores~\cite{ali_hidden_2024}, computed from their general forms derived via SSM, offering an intuitive comparison of their respective roles.

We identify two key properties of ViMs: (1) information is transmitted via \gls{ldsd}, where each hidden state depends on all preceding states and inputs from multiple directions;  and (2) important information is selectively compressed into hidden states during their self-update stages, as Mamba's selective scan recursively encodes the selected features into evolving hidden states. These insights motivate a sequentially-aware approach to token reduction that centers on hidden states. We propose \gls{bnnm} to introduce controllable and self-converging losses within hidden states, and \gls{hsp} to stabilize the fluctuations of hidden states near the class token by leveraging input-dependent parameters for selective compression. Together, BNNM and HSP provide a new perspective on sequential token reduction and the dynamic flow in Mamba, transforming its inherent forgetfulness into a form of stability.
% Our advantages can be summarized as follows: BNNM enables effective information transmission within the sequence by producing a controllable and self-converging loss, paving a feasible way for further adaptation. HSP leverage selective scan in Mamba by retrieving the input-dependent parameters of last layer to achieve a selective information compression into hidden states. Furthermore, BNNM provides a novel sequential token reduction perspective and shed light on the dynamic flow in Mamba, while HSP realizes a minimum perturbation in hidden states and innovatively convert the forgetfulness of Mamba into stability.
We  evaluate our method on ImageNet-1K~\cite{5206848} using ViM-Ti, ViM-S, and ViM-B~\cite{zhu_vision_2024}, achieving consistently high accuracy and strong robustness under aggressive token reduction. Notably, on ViM-Ti, our method keeps the top-1 accuracy drop within 4\% under 40\% token compression, while reducing FLOPs by 22.3\%. Our contributions can be summarized as follows:

1. We identify \gls{ldsd} as the core mechanism of information transmission in ViMs and highlight that information selection and compression occur progressively, requiring consistent maintenance of dynamic representations. These insights prompt us to revisit hidden states and redefine the goal as stabilizing hidden states around the class token.

2. Following the analysis, we propose a novel sequential token reduction pipeline. We introduce BNNM to construct a predictable and controllable loss in the hidden states because BNNM aligns with the original transmission characteristics in ViMs, paving a feasible way for our next step.

3. We leverage selective scan to maintain the online information compression by proposing HSP. HSP minimizes the controllable perturbation loss introduced by BNNM via retrieving previous selective parameters and exploiting the perturbation’s self-converging property.

4. Experiments show our BNNM strategy preserves spatial structure via symmetric token aggregation, significantly mitigating accuracy loss under token reduction. The \gls{hsp} scheme stabilizes hidden states across varying reduction rates, maintaining high accuracy. Our method achieves state-of-the-art performance with the lowest complexity among comparable approaches.

\section{Related work}
\label{gen_inst}
\paragraph{Vision Mamba}

Mamba \cite{gu_mamba_2024} has been shown to be a promising alternative \cite{dao_transformers_2024} to Transformer \cite{vaswani_attention_2017} with only linear complexity of tokens. Recently, abundant works explore the effectiveness of Mamba in computer vision \cite{chen_rsmamba_2024,guo_mambair_2025,hatamizadeh_mambavision_2025,huang_localmamba_2024,li_videomamba_2025,liu_vmamba_2024,patro_simba_2024,pei_efficientvmamba_2025,ruan_vm-unet_2024,shi_multi-scale_2024,yang_plainmamba_2024}, represented by ViMs \cite{zhu_vision_2024}, which focus on bidirectional token scanning. However, these works mainly contribute to backbone mechanism design; further optimization remains still unexplored. Our proposal is an effective inference acceleration method with negligible overhead from a novel sequential token merging perspective.

\paragraph{Token Reduction}

Token reduction is a successful strategy in model compression to balance the computational load and enhance efficiency by reducing amount of the tokens processed. It is popular in efficient Transformers \cite{bolya_token_2023,meng2022adavit,rao2021dynamicvit,yin2022vit} because it doesn’t require specific hardware design nor additional weights \cite{zhan_exploring_2024}, but its  potential hasn’t been fully recognized in Mamba. Several informative works have been done towards this direction \cite{shen_famba-v_2024,shi_faster_2024,zhan_exploring_2024,zhan_rethinking_2024}.  Famba-V \cite{shen_famba-v_2024} is the first who transfer ToMe \cite{bolya_token_2023} in ViT to efficient Mamba training, fusing the most similar pairs in a cross-layer  manner. The authors in  \cite{zhan_exploring_2024}  emphasize the importance of token order for performance restoration. They use the decay factor {\small \(\bar{A}\)} to align hidden states and prune tokens based on attention scores across token channels. R-MeeTo \cite{shi_faster_2024} finds that token merging preserves more information than pruning especially in ViM. It also follows ToMe \cite{bolya_token_2023} by merging token pairs with the nearest distance, relying on a brief retraining stage to restore accuracy. However, these previously successful methods in ViTs prove ineffective when applied to ViMs, primarily due to the distinct and unidentified information flow mechanism in Mamba, particularly regarding token behavior and their interaction with hidden states. 

\section{Revisiting Hidden States}
\subsection{Vision Mamba}
State Space Models (SSMs) map an input sequence \(x(t) \in \mathbb{R}^L \) to an output sequence \( y(t) \in \mathbb{R}^L\) via a hidden state \(h(t) \in \mathbb{R}^N \), with dynamics defined as:

\begin{equation}
\begin{aligned}
    &\hfill h'(t) = \boldsymbol{A}h(t) + \boldsymbol{B}x(t), \hfill &
    &\hfill y(t) = \boldsymbol{C}h(t). \hfill
\end{aligned}
\label{eq:continuous-SSM}
\end{equation}

Mamba is derived from continuous SSMs by discretizing with a timescale \(\Delta\) and the zero-order hold (ZOH) rule (equation~\eqref{eq:discretization}), converting the continuous differential equations (\ref{eq:continuous-SSM}) into the linear recurrence (\ref{eq:discrete-SSM}), and further into the general form (\ref{eq:hidden-states}).

\begin{equation}
\begin{aligned}
     {\bar{A}} = \exp(\Delta \boldsymbol{A}), \quad\quad
 {\bar{B}} = (\Delta \boldsymbol{A})^{-1}(\exp(\Delta \boldsymbol{A}) - \boldsymbol{I}) \cdot \Delta \boldsymbol{B}.  \\
\end{aligned}
\label{eq:discretization}
\end{equation}

\begin{equation}
\begin{aligned}
    &\hfill h_t = {\bar{A_t}}h_{t-1} + {\bar{B_t}}x_{t},\hfill &
    &\hfill  y_t = \boldsymbol{C_t}h_t.   \hfill
\end{aligned}
\label{eq:discrete-SSM}
\end{equation}
\begin{equation}
h_t = \sum\limits_{j = 1}^{t}(\prod
_{k = j+1}^{t}{\bar{A}_k})\bar{B}_jx_j
\label{eq:hidden-states}
\end{equation}

 \paragraph{Hidden Attention in Mamba} 
 \cite{ali_hidden_2024} shows that SSM-based models can be interpreted as attention-driven, where hidden states function can be interpreted as an attention mechanism under suitable transformations with query, key, and value formulations. Specifically, they derive the attention matrices in selective state spaces layers generally as \(\tilde\alpha _{i,j}\) in (\ref{eq:attention})

\begin{equation}
\tilde\alpha _{i,j} = C_i(\prod\limits_{k = j+1}^{i}{\bar{A}_k})\bar{B}_j
\label{eq:attention}
\end{equation}

\paragraph{The pipeline of Vision Mamba}  ViM \cite{zhu_vision_2024} projects flattened 2D patches \(t_p\) using a learnable matrix \(W\), with \(t_{cls}\) denoting the class token representing the entire sequence. We use the middle class token, and \(T_{l-1}\) denotes the input token sequence for the \(l^{th}\) ViM encoder layer.
\begin{equation}
T_0 = [ t^1_pW; t^2_pW; \ldots ;t_{cls};\ldots;  t^J_pW] + Epos
\label{eq:vim}
\end{equation}

\subsection{Mamba Hidden States Analysis and Discovery}
Mamba replaces multi-head attention~\cite{han_survey_2023} with hidden states, prompting us to revisit their role for a more adaptive token reduction strategy. This section presents our analysis of hidden states in ViMs, introducing \gls{ldsd} in~\ref{3.2.1} to characterize information flow between adjacent hidden states, and showing in \ref{3.2.2} that Mamba performs online sensing and compression of important information.

\subsubsection{Theoretical Exploration: Limited Directional Sequential Dependence}
\label{3.2.1}

Hidden states in Mambas and multi-head attention in Transformers represent fundamentally different mechanisms, despite serving similar objectives. While prior works have explored their theoretical differences and potential connections \cite{ali_hidden_2024, han_demystify_2024}, and even suggested structural similarities \cite{dao_transformers_2024}, these insights offer limited practical guidance for token reduction in ViMs. This is mainly due to the absence of a comparative framework that explicitly focuses on the mechanisms of information transmission in ViMs, particularly how tokens and hidden states interact with sequence structures.
We introduce \glsentryfull{ldsd} to address this gap. It captures a distinctive property of hidden states in Mamba-based architectures: each hidden state exhibits localized and directional dependence on the entire sequence, in contrast to the uniform pairwise interactions in Transformers. LDSD provides a new lens for understanding token-sequence relationships in ViMs, which is both theoretically grounded and practically useful, forming the basis for our token merging strategy detailed in Section~\ref{method}.
Concretely, we define LDSD as a structured hidden state evolution process, where a focal hidden state \( h' \) is formed by aggregating directionally transformed previous hidden states \( h \), with each transformation influenced by its corresponding input \( x \). This captures both the temporal sequentiality and spatial directionality unique to Mamba models.  Formally, this dependence can be written as:

\begin{equation}
  \label{eq:LDSD}
  h' = \sum\limits_{Directions,h} SSM(h,x)
\end{equation}

Table \ref{table 1} illustrates the two key components of LDSD: the directional information flow and the core assumption that all input tokens are part of a stream. This implies strong dependencies between each token and the entire sequence, contributing jointly to the final representation. For clarity, we contrast these properties with those of ViTs.
\begin{table}
  \caption{Detailed Analysis about Limited Directional Sequential Dependence}
  \label{table 1}
  \centering
  \begin{tabular}{llll}
    \toprule

    Perspective     & Description     & ViMs & ViTs  \\
    \midrule
    \multirow{3}{*}{Information flow} 
    &Information mechanism & Hidden states  & Multi-head attention   \\
    &Information direction &  Directional & Non-directional    \\
    &Memory ability    & Forgetfulness      & Strong memory \\
    \cmidrule(lr){1-4}
    \multirow{3}{*}{Token assumption} 
    &Logic behind tokens &Sequentialized & Atomized \\
    &Dependence on sequence&Strong dependence&Weak relation \\
    &Complexity&\(O(n)\)&\(O(n^2)\)  \\
    
    \bottomrule
  \end{tabular}
\end{table}
\paragraph{Information flow} Hidden states in ViMs provide a unique flow of sequential information and enable linear-time inference. This process can be compared to a chain of individuals passing information, where each person (i.e., hidden state) refines the message before forwarding it to the next person. In bidirectional or multi-directional Mamba models, multiple such chains exist, passing information in parallel. Unlike ViTs, which rely on computationally intensive attention mechanisms, ViMs aggregate patch information into a single class token through the hidden state chains, resulting in sub-quadratic inference complexity. However, this efficiency introduces trade-offs. ViMs tend to forget early inputs due to the sequential nature of information flow, and the connection between the class token and individual image patches is indirect.
We use the term "Limited Directional" to describe this indirectness, because the class token perceives the entire image only through constrained directional branches formed by hidden state propagation. %a directional flow of information—like looking through constrained branches formed by hidden state chains. 
This makes Mamba’s perceptual system inherently limited and potentially fragile. While some recent works attempt to restore global information by introducing multi-directional flows \cite{liu_vmamba_2024,ma_rs3mamba_2024,zhao_rs-mamba_2024}, we argue that these approaches only partially address the issue. Without an effective coordination mechanism, the directional biases from different scans remain incoherent  \cite{ramachandran_ouromamba_2025}. Even with coordination, the directions remain fundamentally limited.

\paragraph{Sequentialized Token Assumption.} The SSM formulation~\eqref{eq:discrete-SSM} implicitly assumes that tokens are sequentialized, as each input \(x\) updates the hidden state between adjacent time steps. This reflects Mamba’s assumption of a strong dependence between individual tokens and their sequence, as described in~\eqref{eq:LDSD}. This assumption also makes token reduction not a straightforward task because reducing tokens may yield a tricky perturbation to the information chain. We figure out the position of tokens plays an equivalent vital role in forming the information flow as well as the information of tokens itself due to the hidden state chains. For example, if we exchange two arbitrary tokens' positions without changing their values, the output is still interfered because the pretrained parameters of different positions in hidden states only adapt to the inputs from the corresponding positions. Even worse, this perturbation may be pronounced due to the aggregation nature of the chains because each hidden state has to resort to its adjacent states and has no other means to calibrate the information it receives. Therefore, when reducing tokens, we not only need to determine the information conveyed by each token but also recognize the global effect each token gives to the chain of hidden states. 

In essence, ViMs rely on the effective transmission of semantic information through hidden state chains and sequentialized tokens. This insight prompts a rethinking of token reduction in ViMs, where preserving the class token’s hidden state becomes the key objective since all token information is aggregated there. Guided by the principle of \gls{ldsd}, we propose \gls{bnnm} in \ref{BNNM}.

\subsubsection{Experimental Exploration: Online Information Selective Compression }
\label{3.2.2}

\begin{figure}
  \centering  \includegraphics[width=1.0\linewidth]{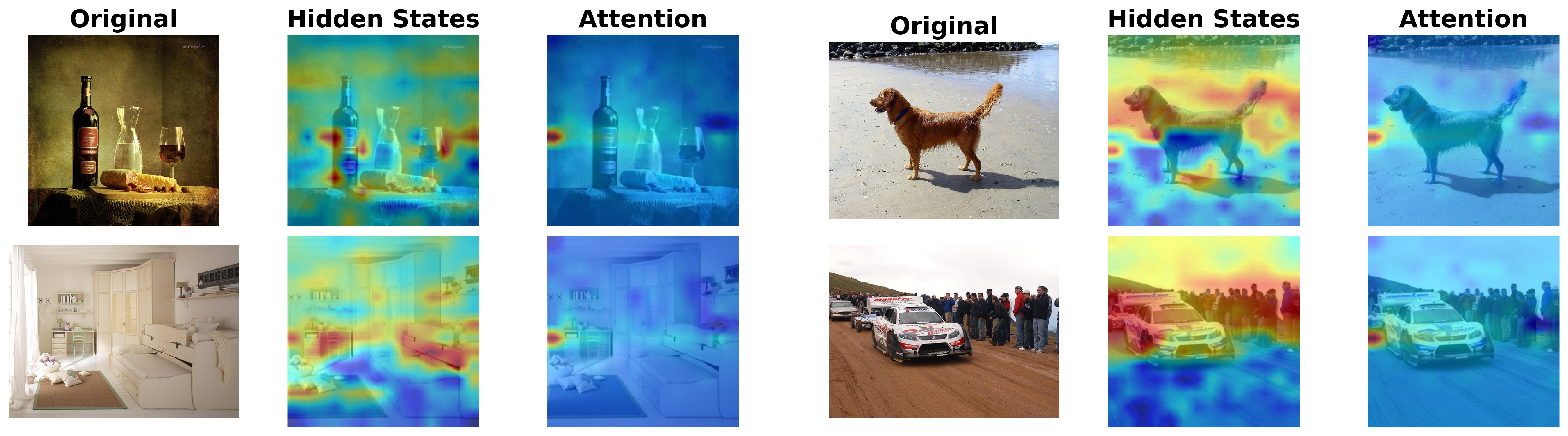}
  \caption{Hidden states show much more semantic information than attention. This phenomenon originates from the selective scan in Mamba.}
  \label{fig1}
\end{figure}
Selective scan plays a dominant role in selecting and compressing the important information from tokens into hidden states \cite{gu_mamba_2024}, and this process happens during the self-update stages of hidden states.

\paragraph{Dominant status of selective scan.}
\label{dominant status}
The success of Mamba is largely credited to selective scan. In brief, selective scan sets that \(B\), \(C\), and \( \Delta \) are all liner projections of input \(x\), with \(A\) and \(B\) being further discretized with \( \Delta \). The function of selective scan in ViMs is intuitively reflected in Figure~\ref{fig1}.
Heat maps of hidden states reveal significantly more semantic and input-dependent information than those of attention matrices. In contrast to the attention formulation in equation~\eqref{eq:attention}, where the input tokens \(x\) are decoupled from the update process, the hidden states \(h\) are updated through a formulation that fuses the input \(x\), as shown in equation~\eqref{eq:hidden-states}. This fundamental difference accounts for the superior semantic expressiveness of hidden states, as illustrated in Figure~\ref{fig1}. The key reason lies in the selective scan mechanism: to fully activate this mechanism, the input must be integrated into the hidden states. This recursive process enables dynamic and input-variant representations, as \(x\) is first projected into {\small \(\bar{A},\bar{B}\)}, which in turn selectively extract features from \(x\). Nevertheless, heat maps of attention consistently exhibit the highest scores around the class token regardless of inputs. This phenomenon stems from the isolation of \(x\) in the attention formulation~\eqref{eq:attention} ,  where the decay factor {\small \(\bar{A}\)} dominates the computation. In our analysis, we refer to  {\small \(\bar{A},\bar{B}\)} as the core components of the selective scan mechanism and omit { \small \(C\)} during inference, as we focus solely on the hidden states and do not require intermediate outputs.

\paragraph{Online compression property.}
\label{online compression property}
The online compression property of SSMs originates from their finite-state nature \cite{gu_mamba_2024}, which inherently compresses information within the hidden states. Unlike attention-based models, SSMs operate without global context (e.g., key-value caches) during inference. They process input tokens asynchronously,  progressively selecting and compressing information via recurrent updates between adjacent hidden states. In contrast, ViTs perform token selection and compression in a synchronous manner by computing attention scores across all tokens simultaneously. Importantly, no token in an SSM is entirely negligible; any meaningful importance metric should holistically reflect each token’s cumulative influence on hidden state evolution and the input-to-output transformation.

\paragraph{Enlightenment} Our insights from revisiting hidden states highlight the importance of preserving LDSD to maintain effective information transmission. Furthermore, the information of the reduced tokens not fed into Mamba should still be selectively compressed to retain features integrity that contributes to the final output.

\section{Sequential Token Merging}
\label{method}

Based on the enlightenment, we propose a general \glsentryfull{stm} method for ViMs. As illustrated in Figure \ref{fig2}, our \gls{bnnm} preserves LDSD, transforming the uncontrollable loss in traditional token reduction into a controllable perturbation within hidden states. To further minimize information loss, we introduce \gls{hsp} via selective approximation based on our fine-grained analysis of online information selective compression. Finally, by leveraging the self-converging nature of merging loss, STM turns Mamba's forgetfulness into a source of great stability.

\subsection{Bidirectional Nearest Neighbor Merging}
\label{BNNM}
\begin{figure}
  \centering  \includegraphics[width=1.0\linewidth]{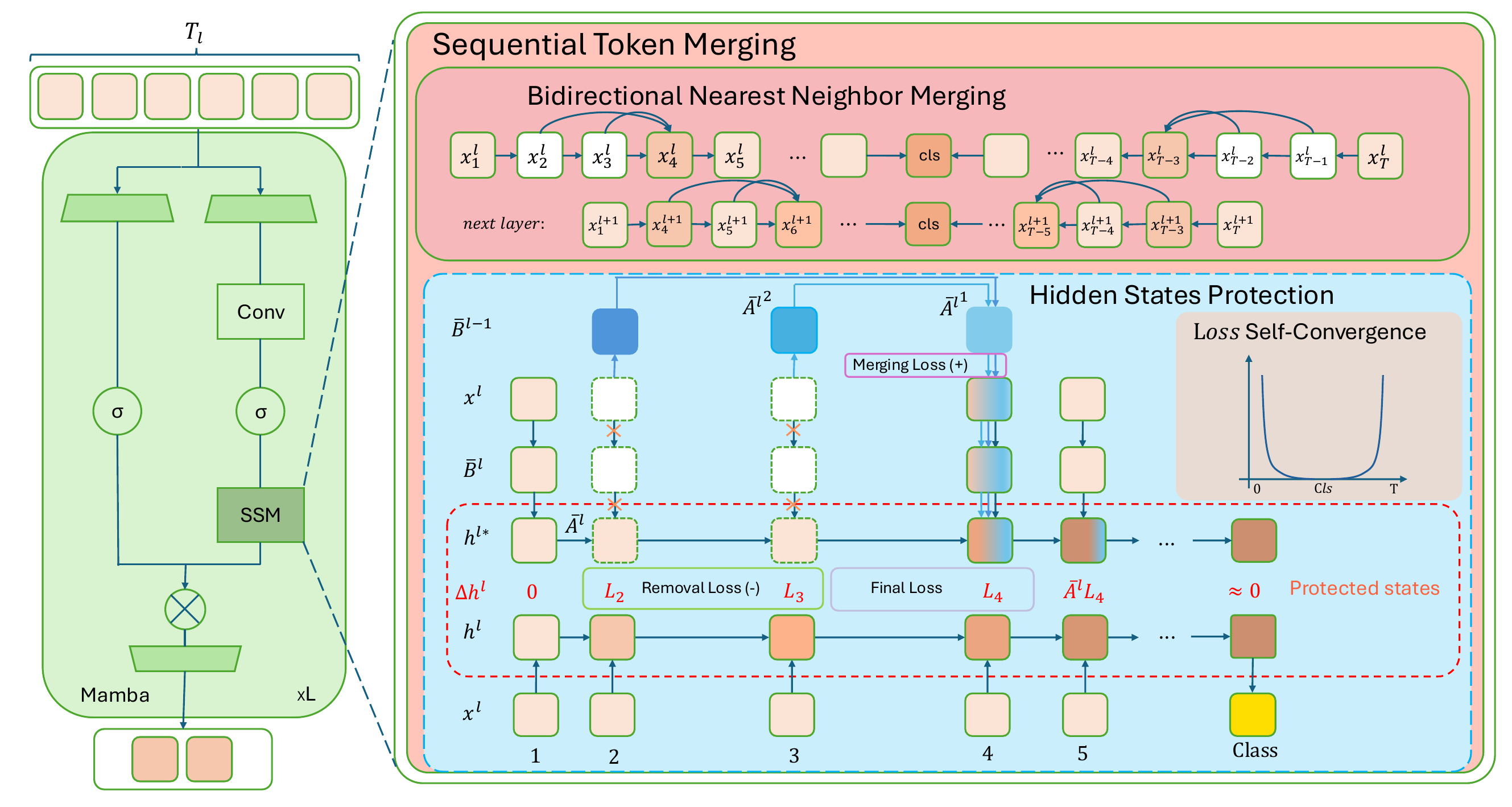}
  \caption{Overview of our proposed Sequential Token Merging (STM) method. It contains two parts: \glsentryfull{bnnm} and \glsentryfull{hsp}. Blue colors indicate tokens with external information, and darker colors indicate tokens with more aggregated information.}
  \label{fig2}
\end{figure}

We propose the \glsentryfull{bnnm} to maintain \gls{ldsd}. At each layer, every reduced token is merged with the nearest remaining neighbor token in its forward direction toward the class token. As illustrated in the \gls{bnnm} component of Figure~\ref{fig2}, this merging process is applied recursively across layers. This approach stabilizes the hidden states and makes perturbations predictable, enabling loss in hidden states to be effectively controlled, as further discussed in Section~\ref{hidden states protection}. Unlike prior methods~\cite{shi_faster_2024,zhan_exploring_2024}, \gls{bnnm} preserves token order and introduces perturbations at predictable locations, making them estimable.
\paragraph{Notation} \label{Notation}We follow the notation system introduced in \cite{zhan_exploring_2024} for our elaboration. In ViM, the input to the \(l^{th}\) layer is a token sequence \(T_{l-1} \in \mathbb{R}^{B\times N\times D}\) (equation~\eqref{eq:vim}), where \(B\) is the batch size, \(N\) is the number of tokens, and \(D\) is the hidden dimension. Each sequence in the batch consists of \(N\) tokens, denoted as\(\{{x_j}\}_{j=0}^{N-1}\). We assume that only \(K\) tokens are retained after merging, while the other \(N - K\) tokens are discarded. Following the notation in \cite{zhan_exploring_2024}, we represent the indices of the retained tokens as \( \{q_k\}_{k=0}^{K-1}\),  \(q_s < q_t\) for all \(s < t\). To make the problem more tractable, consider two adjacent remaining tokens \(x_{q_k}\) and \(x_{q_{k+1}}\). If \(q_{k+1} - q_k > 1\), this indicates that some tokens have been reduced between them. Let \(R_k \geq 1\) denote the number of reduced tokens between \(q_{k}\) and \(q_{k+1}\), whose indices are given by \(\{{q_k + i}\}_{i=1}^{R_k}\). These indices satisfy \(q_{k} < q_{k} + 1 < \dots < q_{k} + R_k < q_{{k+1}}\). To emphasize the distinction between \(q_{k+1}\) and \(q_k + 1\), we adopt the convention of using round brackets in expressions such as \(q_{(k+1)}\) and \(q_{(k)} + 1\), though their meanings remain unchanged. We use \( h^*\) to denote the modified hidden states, and we use \(h\) to denote the original hidden states.

\paragraph{Removal Loss and Merging Loss.} The removal loss measures the change in hidden states caused solely by removing a token at its original position \(q_{(k)}+i\), capturing the information loss from that token. In contrast, the merging loss quantifies the perturbation introduced at the merging position \(q_{(k+1)}\) due to incorporating additional information from the merged token. Because hidden states are updated sequentially, the merging loss cannot be directly measured at \(q_{(k+1)}\) without including accumulated effects. Thus, we define it as the final loss at \(q_{(k+1)}\) minus the propagated removal losses from all merged tokens. Specifically, we denote the merging loss as {\small \(L_{q_{(k+1)_M}} \)} in equation~\eqref{eq:merging loss}, in contrast to the overall loss {\small \(L_{q_{(k+1)}} \)}~\eqref{eq:final loss}. The removal loss originally caused by discarding a reduced token at position \(q_{(k)}+i\), and then propagated to \(q_{(k+1)}\), is denoted as {\small \(\text{SSM}(L_{q_{(k)}+i})\)}. This formulation reflects how removing a token affects subsequent hidden states through sequential dependencies, as modeled by the SSM. Given that both the removed position \(q_{(k)}+i\) and the merging target position \(q_{(k+1)}\) are predetermined, we can explicitly compute these two types of loss without introducing other effects. This controllability ensures that \gls{bnnm} provides a stable and predictable loss behavior, forming a reliable foundation for further loss minimization in later stages, especially during selective-aware token merging (Section~\ref{token merging}).
\begin{align}
L_{q_{(k)}+i} &= h_{q_{(k)}+i} - h^*_{q_{(k)}+i} \quad (\text{Removal Loss})\label{eq:removal loss} \\
L_{q_{(k+1)}} &= \sum_{\text{Directions}} (h_{q_{(k+1)}} - h^*_{q_{(k+1)}})  \quad (\text{Final Loss})  \label{eq:final loss}\\
 L_{q_{(k+1)_M}} &= L_{q_{(k+1)}}
- \sum_{\text{Directions}, i} \text{SSM}(L_{q_{(k)}+i}) \quad (\text{Merging Loss}) \label{eq:merging loss}
\end{align}
The following section details how we leverage the selective scan mechanism to estimate and compress the information of reduced tokens into hidden states, even without passing through Mamba. Our method strategically employs the merging loss to compensate for the removal loss.

\subsection{ Hidden States Protection}
\label{hidden states protection}
To align with Mamba’s online compression property (Section~\ref{3.2.2}), we propose selective-aware token merging, which aims to minimize the final hidden state loss as defined in equation~\eqref{eq:final loss}. Additionally, we transform the forgetfulness of Mamba into stability against perturbations.
\subsubsection{Selective-aware token merging}
\label{token merging}
The general formulation of our token merging strategy is presented below. It is derived by comparing the hidden states before and after token removal to obtain the compensation term, with detailed derivations provided in Appendix~\ref{appendix: mergefwd}.
\begin{align}
\text{Merge}_{fwd}(x_{q_{(k-1)}+1},x_{q_{(k-1)}+2},\dots,x_{q_{(k)}}) 
&= \underbrace{\sum\limits_{j = 1}^{R_{k-1}} (\prod\limits_{n=j+1}^{q_{(k)}}{\bar{A}_n} )\underbrace{\frac{\bar{B}^{(l-1)}_{q_{(k-1)}+j}}{\bar{B}^{(l-1)}_{q_{(k)}}} x_{q_{(k-1)}+j}}_{\text{Removal loss estimation (-)}}}_{\text{Merging loss compensation (+)}} + \underbrace{x_{q_{(k)}}}_{\text{Original token}}
\label{eq: merge_fwd}
\end{align}
\paragraph{Removal loss estimation (-) \& Merging loss compensation (+).}  To incorporate token selectiveness into the merging process, we begin by estimating the removal loss using the previous layer’s \(\bar{B}^{(l-1)}\) terms corresponding to the reduced token at \(q_{(k-1)}+j\) and the remaining token at \(q_{(k)}\). This estimated hidden state loss is then propagated through the current layer to compute the merging loss compensation, i.e., each term in the forward merging function is scaled by an exponentially decaying factor constructed from multiple \(\bar{A}\) values spanning from the reduced position to the merging point. Therefore, our approach leverages the dominant status of the selective scan mechanism (Section~\ref{dominant status}) in Mamba. By integrating {\small\(\bar{A},\bar{B}\)} into the merging strategy, it fully utilizes Mamba’s inherent selectivity.

\subsubsection{From Forgetfulness to Stability: The Self-Convergence of Loss}
\paragraph{Corollary (Distance Fading Rule):} 
\begin{equation}
L_{q_{(k+1)}+1} = \bar{A}_{q_{(k+1)}+1}L_{q_{(k+1)}}  
\label{eq:corollary 1 fading property}
\end{equation}
\begin{equation}
L_{cls} = \bar{A}^{|cls - p|}L_{p}
\label{eq:corollary 2 distance fading rule}
\end{equation}

\begin{figure}
  \centering  \includegraphics[width=1.0\linewidth]{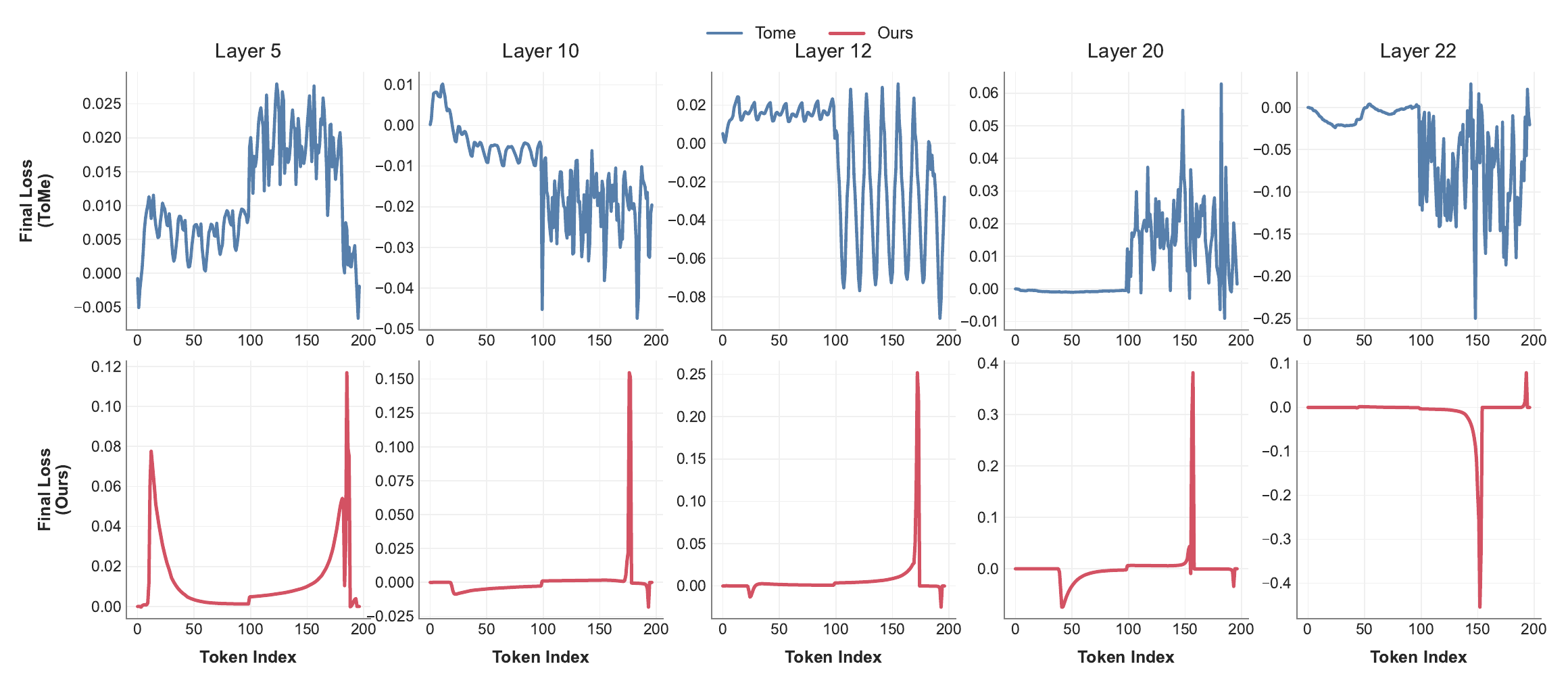}
  \caption{This figure presents a layer-wise comparison of the final loss introduced by ToMe-based methods in ViM \cite{bolya_token_2023}\cite{shen_famba-v_2024}\cite{shi_faster_2024}, and our proposed method (STM). As illustrated, the  merging compensation in STM ensures that hidden states undergo only minimal perturbations at merging points and rapidly converge to zero, stabilizing the hidden states around the class token.}
  \label{fig 3}
\end{figure}

After compensation, we can obtain the final loss with equation~\eqref{eq:final loss} and derive the corollary.
Due to the linear property of Mamba, the final loss at the merging position {\small \(L_{q_{(k+1)}} \)} decays exponentially over time, with the hidden state difference at the next token position scaled by {\small \(\bar{A}_{q_{(k+1)}+1}\)}. The proof is provided in Appendix \ref{proof for corollary 1}. Combining this property with the bidirectional Mamba architecture, we calculate the ultimate impact of a token merged at position \(p\) on the class token. Using the middle class token in the ViM setting as an example, we derive the distance fading rule~\eqref{eq:corollary 2 distance fading rule}, where \(cls\) denotes the position of the class token. For simplicity, we replace the product {\small \(\prod_{j = p+1}^{cls}{\bar{A_j}}\)} with {\small\(\bar{A}^{|cls - p|}\)}, since the selective nature of {\small \(\bar{A}\) }arises from \( \Delta\) \cite{gu_mamba_2024}, making \( \bar{A}\) approximately a constant factor. This is also supported by the exponential degradation curves shown in Figure \ref{fig 3}.
This formulation highlights how the distance between the merged token and the class token effects the final hidden representation. Importantly, this property proves useful in our experiments. Corollary~\eqref{eq:corollary 2 distance fading rule} finally motivates a layer-wise merging strategy, where tokens are merged progressively from distant to closer positions, thereby leveraging the loss’s self-convergence. As a result, our method facilitates a truly data-free design, meaning it doesn't rely on any data for fine-tuning or retraining the compressed model.

\section{Experiment Results}
\label{experiment results}
We implement our method using PyTorch \cite{paszke_pytorch_2019} for scientific computation. To validate both our theoretical analyses and the empirical effectiveness of STM, extensive image classification experiments are conducted on the ImageNet-1K \cite{5206848} validation dataset, using 4 NVIDIA RTX 4090 GPUs.  We conduct a comparison against three state-of-the-art token reduction techniques in ViM: Token Recognition \cite{liang_not_2022}, Hidden State Alignment \cite{ zhan_exploring_2024}, and R-MeeTo \cite{shi_faster_2024}.  To enable a fair comparison and demonstrate our super performance, the result of \cite{zhan_exploring_2024} and \cite{shi_faster_2024} are yielded directly after token reduction without fine-tuning nor retraining, while scores of \cite{liang_not_2022} are the fine-tuned results reported in \cite{zhan_exploring_2024}. For the bidirectional ViM-Ti, ViM-S, and ViM-B \cite{zhu_vision_2024} models, we apply a uniform token reduction rate of 20\% across all settings. STM consistently applies forward merging function~\eqref{eq: merge_fwd} to merge two tokens in each direction at every layer throughout the model. ViM-Ti validation takes under 2 hours, ViM-S around 4 hours, and ViM-B around 6 hours. Table \ref{tab:comparison} provides ImageNet validation top-1 accuracies and the accompanying FLOPs. %The variation of performance of ViM-Ti and ViM-S under different token reduction rate is listed in Table \ref{tab:ablation}.
\medskip
\begin{table}[h]
\centering
\caption{Comparison of different methods on top-1 accuracy and FLOPs. \textbf{Bold} is the best, and \underline{underline} is the second best.}
\label{tab:comparison}
\resizebox{\textwidth}{!}{%
\begin{tabular}{lcccccc cccccc}
\toprule

\multirow{2}{*}{\textbf{Method}}& \multicolumn{6}{c}{\textbf{top-1 acc. (\%)}} & \multicolumn{6}{c}{\textbf{FLOPs (G)}} \\
& \multicolumn{2}{c}{\textbf{Vim-Ti}} &
\multicolumn{2}{c}{\textbf{Vim-S}} &
\multicolumn{2}{c}{\textbf{Vim-B}}&\multicolumn{2}{c}{\textbf{Vim-Ti}} &
\multicolumn{2}{c}{\textbf{Vim-S}} &
\multicolumn{2}{c}{\textbf{Vim-B}}  \\

\midrule
Vim (Baseline) & 76.1 & 0.0 & 80.5 & 0.0 & 81.9 & 0.0 & 1.45 & 0.00 & 5.08 & 0.00 & 18.87 & 0.00 \\
Token Recognition & $\underline{71.3}$ & $\underline{4.8}\downarrow$ & $74.8$ & $5.7\downarrow$ & -- & -- & $\underline{1.29}$ & $\underline{0.16}\downarrow$ & $\boldsymbol{3.57}$ & $\boldsymbol{1.51}\downarrow$ & -- & -- \\
Hidden State Alignment & $49.3$& $26.8\downarrow$& $72.0$& $8.5\downarrow$& $\underline{79.6}$& $\underline{2.3}\downarrow$& 1.29& $0.16\downarrow$ & $4.48$& $0.60\downarrow$& $\underline{16.58}$&$\underline{2.29}\downarrow$\\
R-MeeTo & $52.3$ & $23.8\downarrow$ & $\underline{78.7}$ & $\underline{1.8}\downarrow$ & $79.1$ & $2.8\downarrow$ & $1.41$ & $0.04\downarrow$ & $4.73$ & $0.35\downarrow$ & $17.97$& $0.9\downarrow$\\
\rowcolor{lightblue}
Ours & $\boldsymbol{75.1}$& $\boldsymbol{1.0}\downarrow$& $\boldsymbol{79.3}$ & $\boldsymbol{1.2}\downarrow$ & $\boldsymbol{79.8}$& $\boldsymbol{2.1} \downarrow$& $\boldsymbol{1.28}$& $\boldsymbol{0.17}\downarrow$& $\underline{4.23}$ & $\underline{0.85}\downarrow$ &$\boldsymbol{15.73}$& $\boldsymbol{3.14}\downarrow$\\
\bottomrule
\end{tabular}
}
\end{table}
\medskip
\paragraph{Results} As shown in Table \ref{tab:comparison}, our method consistently achieves the highest accuracy among all compared approaches across all models. ViM-Ti only has a 1.0\% drop in accuracy compared to the unreduced baseline, with 25.8\% accuracy higher than Hidden States Alignment and 22.8\% higher than R-MeeTo. For ViM-S, our method achieves 5.6\% higher performance restoration rate than Token Recognition because Token Recognition prunes tokens excessively. For ViM-B, our method continues to outperform other approaches with wide margins in accuracy, along with a 16\% FLOPs reduction compared to 12.1\% of Hidden States Alignment and 4.7\% of R-MeeTo respectively.

\paragraph{Complexity Comparison}
\label{complexity}
Despite achieving the highest accuracy across all models, our method requires significantly fewer FLOPs than Hidden State Alignment and R-MeeTo. This efficiency is attributed to the uniform merging strategy inherited from BNNM, which avoids the costly computation of pairwise distances or importance scores for token selection. These computations grow quadratically with the number of tokens and hidden dimensions, leading to substantial overhead in larger models. In contrast, our method maintains a lightweight inference process. Experiments validate that this simple yet principled merging approach offers a strong balance between efficiency and accuracy.
% \paragraph{Complexity Comparison} Given our highest accuracy across all models, our method still requires much smaller FLOPs than Hidden State Alignment and R-MeeTo. This is because we perform a uniform merging mechanism inherited from BNNM, enabling a faster inference with no need to compute distance nor importance scores to detect tokens, thus saving us a large amount of computation. This phenomenon is pronounced in models with greater parameters when the  computation cost of distance and importance scores gets more expensive with models with higher dimensions. The experiments demonstrate our method is more effective and efficient even we simply conduct merging in a uniform way with proper merging method.

\subsection{Ablation \& Analysis}
\paragraph{Both sides v.s. One side} In Table \ref{tab:ablation}, we compare the impact of merging tokens on one side versus both sides symmetrically in bidirectional Mamba. STM merges 1, 2, 2, and 2 tokens per direction at each layer to achieve token reductions of 15\%, 20\%, 30\%, and 40\%, respectively. On ViM-Ti, the both-sides merging strategy reduces degradation from 
45.0\%\(\downarrow\) to 1.0\%\(\downarrow\) at 15\% token reduction. This results demonstrate the effectiveness of our bidirectional merging, which better preserves spatial structure by symmetrically aggregating contextual information with minimal perturbation. On ViM-S, the difference in accuracy between one-side and both-sides merging is negligible, but one-side merging results in slightly higher FLOPs. Given ViM-S’s larger capacity, it is more resilient to token reduction, with minimal performance degradation. Therefore, we do not include this ablation on ViM-B, as its even larger size shows negligible degradation under our method.

\paragraph{Effect of different reduction rate}
\label{different reduction rate}
Table \ref{tab:ablation} shows that STM maintains strong performance under varying token reduction rates. Notably, STM incurs only minimal performance drops of 3.9\%\(\downarrow\) and 1.4\%\(\downarrow\) under a 40\% token reduction for ViM-Ti and ViM-S respectively, highlighting the stability and effectiveness of our merging strategy.
For a detailed analysis about BNNM, we focus on both-sides merging. ViM-Ti shows a consistent trade-off between accuracy and FLOPs as reduction rate increases. ViM-S, however, peaks in both accuracy and FLOPs at 20\% reduction, with minor accuracy fluctuations and a unique FLOPs rebound due to ViM-S's higher feature dimension, indicating that merging in ViM-S becomes increasingly costly relative to the computation saved.

\medskip
\begin{table}[h]
\centering
\caption{Ablation comparison between one side merging and both sides merging, along with performance under varying token reduction rates. Top-1 acc. (\%) and FLOPs (G) are reported. \textbf{Bold} is the best.}
\label{tab:ablation}
\resizebox{\textwidth}{!}{%
\begin{tabular}{llcccccccc | cccccccc}
\toprule
\multirow{2}{*}{\textbf{Model}} &\multirow{2}{*}{\textbf{Mode}}& \multicolumn{8}{c}{\textbf{Token Merging Rate (\%)}} & \multicolumn{8}{c}{\textbf{FLOPs (G)}} \\
\cmidrule(lr){3-10} \cmidrule(lr){11-18}
&& \multicolumn{2}{c}{\textbf{15\%}} & \multicolumn{2}{c}{\textbf{20\%}} & \multicolumn{2}{c}{\textbf{30\%}} & \multicolumn{2}{c}{\textbf{40\%}}&\multicolumn{2}{c}{\textbf{15\%}}&\multicolumn{2}{c}{\textbf{20\%}} &
\multicolumn{2}{c}{\textbf{30\%}} &
\multicolumn{2}{c}{\textbf{40\%}} \\
\midrule
\multirow{2}{*}{ViM-Ti} & One sides & $31.1$ & $45.0\downarrow$ & $31.1$ & $45.0\downarrow$ & $34.6$ & $41.5\downarrow$ & $34.6$ & $41.5\downarrow$  & $1.29$& $0.16\downarrow$& $1.28$& $0.17\downarrow$& $1.15$& $0.30\downarrow$& $1.12$& $0.33\downarrow$\\

\rowcolor{lightblue} &Both sides & $\boldsymbol{75.1} $ & $\boldsymbol{1.0}\downarrow$ & $75.0$ & $1.1\downarrow$ & $73.3$ & $2.8\downarrow$ & $72.2$ & $3.9\downarrow$  & $1.29$& $0.16\downarrow$& $1.28$& $0.17\downarrow$& $1.15$& $0.30\downarrow$& $\boldsymbol{1.12}$& $\boldsymbol{0.33}\downarrow$\\

\multirow{2}{*}{ViM-S} & One sides & $79.1$& $1.4\downarrow$ & $79.2$ & $1.3\downarrow$ & $79.1$ & $1.4\downarrow$ & $79.1$ & $1.4\downarrow$  & $4.72$& $0.36\downarrow$& $4.24$& $0.84\downarrow$& $4.91$& $0.17\downarrow$& $5.08$& $0.00\downarrow$\\

\rowcolor{lightblue} &Both sides & $79.1$& $1.4\downarrow$ & $\boldsymbol{79.3}$& $\boldsymbol{1.2}\downarrow$& $79.1$ & $1.4\downarrow$&  $79.1$& $1.4\downarrow$& $4.72$& $0.36\downarrow$& $\boldsymbol{4.24}$& $\boldsymbol{0.84}\downarrow$& $4.72$& $0.36\downarrow$& $4.66$& $0.42\downarrow$\\
\bottomrule
\end{tabular}%
}
\end{table}
\medskip

\section{Conclusion and Limitation}
\label{conclustion & limitation}
In this paper, we introduce a novel Sequential Token Merging method for ViMs. To develop a more adaptive token reduction technique, we first analyze the hidden states in Mamba and propose the concept of \gls{ldsd} to describe the unique information transmission mechanism across hidden state chains. We identify the information selection and compression process occurs during interactions between inputs and hidden states. Guided by our insights into hidden states, we apply the \gls{bnnm} strategy to preserve the original dependence relationships and retain essential information through retrieving ViM’s selective scan. By compensating for perturbations and leveraging the self-convergence property of hidden states, we stabilize the key hidden state at the class token position. Our extensive experiments demonstrate the effectiveness of our STM strategy and provide insights into the dynamic information flow in Mamba, offering directions for future research on dynamic mechanisms. Although our method is effective, the inter-layer continuity of {\small \(\bar{B}\)} may not always hold, which limits the accuracy of the current estimation. While our uniform merging strategy is computationally efficient, it can be further optimized to achieve better balance and performance.
% In this paper, we propose a novel sequential token merging method in Vision Mambas. In order to design a more adaptive token reduction method in ViMs, we first analyze the hidden states in Mamba carefully. Then we propose a new concept named limited directional sequential dependence to describe its unique mechanism to transmit information through hidden state chains inclusively. Additionally, we determine the information selection and compression process happens during the interactions between inputs and hidden states. Guided by our discoveries about hidden states in Mamba, we use \gls{bnnm} strategy to align with the original dependence relationship throughout the hidden states. We maintain the important information embedded in input tokens by leveraging ViM's selective scan. Thanks to proper compensation and the self-convergence property of the controllable perturbation in hidden states, we can stabilize the key hidden state at class token position. Our extensive experiments prove the effectiveness of sequential token merging strategy and shed light on the dynamic information flow in Mamba, guiding future research on dynamic mechanisms in Mamba. Though our method is general and stable under high compression rate, the efficiency is limited by baseline.

{
\small

\bibliographystyle{abbrv}
\bibliography{main}
}

%%%%%%%%%%%%%%%%%%%%%%%%%%%%%%%%%%%%%%%%%%%%%%%%%%%%%%%%%%%%

\newpage
\appendix
\input{sec/appendix}

%%%%%%%%%%%%%%%%%%%%%%%%%%%%%%%%%%%%%%%%%%%%%%%%%%%%%%%%%%%%

\newpage

\end{document}

%% file: sec/appendix.tex
\textbf{{\Large Appendix for Mathematical Formulations}}

\section{Merging Process}
\label{appendix: mergefwd}
The forward merging formula is derived as follows.
For clarity, we first extend the hidden state equation~\eqref{eq:hidden-states} at two remaining position \(q_{(k-1)}\) and \(q_{(k)}\):
\begin{align}
\label{eq:hq(k-1)}
h_{q_{(k-1)}} &= \bar{A}_{q_{(k-1)}}\bar{A}_{q_{(k-1)}-1}\ldots\bar{A}_2\bar{B}_1x_1\\
&\quad + \bar{A}_{q_{(k-1)}}\bar{A}_{q_{(k-1)}-1}\ldots\bar{A}_3\bar{B}_2x_2 \nonumber \\
&\quad + \ldots \nonumber\\
&\quad +\bar{A}_{q_{(k-1)}}\bar{B}_{q_{(k-1)}-1}x_{q_{(k-1)}-1}\nonumber \\
&\quad +\bar{B}_{q_{(k-1)}}x_{q_{(k-1)}\nonumber}
\end{align}

\begin{align}
\label{eq:hq(k)}
h_{q_{(k)}} &= \bar{A}_{q_{(k)}}\bar{A}_{q_{(k)}-1}\ldots\bar{A}_{q_{(k-1)}+1}\bar{A}_{q_{(k-1)}}\bar{A}_{q_{(k-1)}-1}\ldots\bar{A}_2\bar{B}_1x_1\\
&\quad +\bar{A}_{q_{(k)}}\bar{A}_{q_{(k)}-1}\ldots\bar{A}_{q_{(k-1)}+1}\bar{A}_{q_{(k-1)}}\bar{A}_{q_{(k-1)}-1}\ldots\bar{A}_3\bar{B}_2x_2 \nonumber\\
&\quad + \ldots \nonumber\\
&\quad +\bar{A}_{q_{(k)}}\bar{A}_{q_{(k)}-1}\ldots\bar{A}_{q_{(k-1)}+1}\bar{A}_{q_{(k-1)}}\bar{B}_{q_{(k-1)}-1}x_{q_{(k-1)}-1}\nonumber\\
&\quad +\bar{A}_{q_{(k)}}\bar{A}_{q_{(k)}-1}\ldots\bar{A}_{q_{(k-1)}+1}\bar{B}_{q_{(k-1)}}x_{q_{(k-1)}}\nonumber\\
&\quad + \bar{A}_{q_{(k)}}\bar{A}_{q_{(k)}-1}\ldots\bar{A}_{q_{(k-1)}+2}\bar{B}_{q_{(k-1)}+1}x_{q_{(k-1)}+1}\nonumber\\
&\quad + \ldots\nonumber\\
&\quad +\bar{A}_{q_{(k)}}\bar{B}_{q_{(k)}-1}x_{q_{(k)}-1} \nonumber\\
&\quad +\bar{B}_{q_{(k)}}x_{q_{(k)}}\nonumber
\end{align}
By examining the two equations, we observe that the second can be expressed in terms of the first as follows,
\begin{align}
\label{hq(k-1)-->hq(k)}
h_{q_{(k)}} &= (\prod_{j = q_{(k-1)}+1}^{q_{(k)}}\bar{A_j}) h_{q_{(k-1)}}\\
&\quad + \bar{A}_{q_{(k)}}\bar{A}_{q_{(k)}-1}\ldots\bar{A}_{q_{(k-1)}+2}\bar{B}_{q_{(k-1)}+1}x_{q_{(k-1)}+1}\nonumber\\
&\quad + \ldots\nonumber\\
&\quad +\bar{A}_{q_{(k)}}\bar{B}_{q_{(k)}-1}x_{q_{(k)}-1} \nonumber\\
&\quad +\bar{B}_{q_{(k)}}x_{q_{(k)}}\nonumber
\end{align}
Now we merge the reduced tokens \(\{{x_{q_{(k-1)}+j}}\}_{j=1}^{R_{k-1}}\) into the nearest remaining neighborhood token \(x_{q_{(k)}}\), and let us introduce a merging function \(x^*_{q_{(k)}} = f( x_{q_{(k-1)}+1},x_{q_{(k-1)}+2},x_{q_{(k-1)}+3}, \dots ,x_{q_{(k)}} )\) to describe the merging process. Aligning with our notation in \ref{Notation}, \(R_{k-1}\) is the number of the reduced tokens between \(x_{q_{(k-1)}}\) and \(x_{q_{(k)}}\). 
We denote the modified hidden state at the merging position \(q_{(k)}\) as \(h^*_{q_{(k)}} \), and it can be computed as follows,
\begin{align}
\label{merging fun --> hq(k)}
h^*_{q_{(k)}}  &= (\prod_{j = q_{(k-1)}+1}^{q_{(k)}}\bar{A_j}) h_{q_{(k-1)}}\\
&\quad + \bar{B}_{q_{(k)}}x^*_{q_{(k)}} \nonumber
\end{align}

To ensure that the new hidden state matches the original \(h^*_{q_{(k)}}\) (equation~\eqref{eq:hq(k)}), we aim to design a merging function that maintains this consistency.
\begin{align}
Objective:x^*_{q_{(k)}} = f( x_{q_{(k-1)}+1},x_{q_{(k-1)}+2},x_{q_{(k-1)}+3}, \ldots,x_{q_{(k)}}) \quad s.t.\quad h^*_{q_{(k)}} = h_{q_{(k)}} 
\end{align} 

By comparing equations~\eqref{hq(k-1)-->hq(k)} and~\eqref{merging fun --> hq(k)}, the differing term can be identified as follows:
\begin{align}
\bar{B}_{q_{(k)}}x^*_{q_{(k)}} &= \bar{A}_{q_{(k)}}\bar{A}_{q_{(k)}-1} \cdots \bar{A}_{q_{(k-1)}+2} \bar{B}_{q_{(k-1)}+1} x_{q_{(k-1)}+1} \label{eq:main} \\
&\quad + \cdots \nonumber \\
&\quad + \bar{A}_{q_{(k)}} \bar{B}_{q_{(k)}-1} x_{q_{(k)}-1} \nonumber \\
&\quad + \bar{B}_{q_{(k)}} x_{q_{(k)}} \nonumber
\end{align}
Solving the equation yields:
\begin{align}
x^*_{q_{(k)}} &= \bar{A}_{q_{(k)}}\bar{A}_{q_{(k)}-1}\ldots\bar{A}_{q_{(k-1)}+2}\frac{\bar{B}_{q_{(k-1)}+1}}{\bar{B}_{q_{(k)}}}x_{q_{(k-1)}}\\
&\quad + \ldots\nonumber\\
&\quad +\bar{A}_{q_{(k)}}\frac{\bar{B}_{q_{(k)}-1}}{\bar{B}_{q_{(k)}}}x_{q_{(k)}-1} \nonumber\\
&\quad +x_{q_{(k)}}\nonumber\\
&= \sum_{j = q_{(k-1)}+1}^{q_{(k)}}(\prod_{n =j+1}^{q_{(k)}}\bar{A}_n)\bar{B}_jx_j\nonumber
\end{align}
Rewrite the summation notation, we obtain:
\begin{align}
x^*_{q_{(k)}} &= f( x_{q_{(k-1)}+1},x_{q_{(k-1)}+2},x_{q_{(k-1)}+3}, \ldots,x_{q_{(k)}})\\
&=  \sum_{j = 1}^{q_{(k)}-q_{(k-1)}}(\prod_{n =j+1}^{q_{(k)}}\bar{A}_n)\frac{\bar{B}_{q_{(k-1)}+j}}{\bar{B}_{q_{(k)}}}x_{q_{(k-1)}+j} \nonumber\\
&= \sum_{j = 1}^{R_{k-1}}(\prod_{n =j+1}^{q_{(k)}}\bar{A}_n)\frac{\bar{B}_{q_{(k-1)}+j}}{\bar{B}_{q_{(k)}}}x_{q_{(k-1)}+j} + x_{q_{(k)}}\nonumber
\end{align}
The last term is exactly the same as~\eqref{eq: merge_fwd}, except for the notation of \((l-1)\) of \(\bar{B}\). Because the selective mechanism of Mamba, we cannot compute \(\bar{B}^{(l)}\) if we reduce the corresponding token \(x\) at the layer it's reduced. So we have to refer to the \(\bar{B}^{(l-1)}\) in the last layer to estimate \(\bar{B}^{(l)}\). This is where almost all of loss come from.
\section{Proof of the Corollary}
\label{proof for corollary 1}
We leverage the linearity of Mamba to derive the distance fading rule of the loss. According to~\eqref{eq:final loss}, the final loss is defined as:
\begin{align}
L_{q_{(k+1)}} &=  h_{q_{(k+1)}} - h^*_{q_{(k+1)}}\\
L_{q_{(k+1)}+1} &=  h_{q_{(k+1)}+1} - h^*_{q_{(k+1)}+1}\nonumber
\end{align}
Then, based on the discrete SSM formulation in equation~\eqref{eq:discrete-SSM}, we have:
\begin{equation}
h^*_{q_{(k+1)}+1} = \bar{A}_{q_{(k+1)}+1} h^*_{q_{(k+1)}} + \bar{B}_{q_{(k+1)}+1}x_{q_{(k+1)}+1}\\
\end{equation}
\begin{equation}
h_{q_{(k+1)}+1} = \bar{A}_{q_{(k+1)}+1} h_{q_{(k+1)}} + \bar{B}_{q_{(k+1)}+1}x_{q_{(k+1)}+1}\nonumber
\end{equation}
So we get,
\begin{align}
L_{q_{(k+1)}+1} &=  \bar{A}_{q_{(k+1)}+1}[h_{q_{(k+1)}} - h^*_{q_{(k+1)}}]\\
 &=  \bar{A}_{q_{(k+1)}+1}L_{q_{(k+1)}}\nonumber
\end{align}
This completes the derivation of the corollary.